# Investigation of the condominium building collapse in Surfside, Florida: A video feature tracking approach

Xiangxiong Kong

Department of Physics and Engineering Science, Coastal Carolina University

P.O. Box 261954, Conway, SC, USA  29528-6054

Danny Smyl

Department of Civil, Coastal, and Environmental Engineering, University of South Alabama

150 Student Services Drive, Shelby Hall 3142, Mobile, AL, USA 36688

Abstract

On June 24, 2021, a 12-story condominium building (Champlain Towers South) in Surfside, Florida partially collapsed, resulting in one of the deadliest building collapses in United States history with 98 people confirmed deceased. In this work, we analyze the collapse event using a video clip that is publicly available on social media. In our analysis, we apply computer vision algorithms to corroborate new information from the video clip that may not be readily interpreted by human eyes. By comparing the differential features against different video frames, our proposed method is used to quantify the falling structural components by mapping the directions and magnitudes of their movements. We demonstrate the potential of this video processing methodology in investigations of catastrophic structural failures and hope our approach may serve as a basis for further investigations into structure collapse events.

# 1. Introduction

Structural engineering is critical to our society for enabling safe, efficient, and economical designs to resist the gravity load and loads caused by natural or man-made disasters. Despite significant efforts made in improving the structural design, catastrophic structural failures remain an unfortunate reality [1, 2, 3, 4, 5, 6]. The most recent such event in the United States is the partial collapse of the 12-story condominium building (Champlain Towers South) in Surfside, Florida on June 24, 2021 [7]. This is one of the deadliest building collapses in American history with 98 people confirmed deceased as of July 22, 2021.

To date, extensive efforts have been made into studying the failure mechanisms of civil structures. One engineering approach is to numerically simulate the structure collapse behavior against extreme structural loads (e.g., strong ground motions, extreme blast loads) through finite element methods [8, 9, 10]. As a result, outputs such as stresses, strains, force, and deformations of the structural members as well as floor accelerations and story drifts can be generated from the numerical model. For instance, Xu et al. [8] investigated the internal force redistribution by simulating the progressive collapse of a dome structure. The changes in strains of truss members and displacements of joints were illustrated after critical structural members were removed from the numerical model. Nevertheless, numerical simulation methods are prone to error [11] due to the uncertainties in determining multiple model constraints such as nonlinear properties of the engineering materials [12], lack of consideration of geometric imperfections [8, 9], and convergence issues in computational algorithms [10].

Alternatively, performing experimental tests [12, 13, 14, 15] in the structural laboratories can obtain a more comprehensive understanding of the structure collapse by directly taking measurements of the critical locations of structures. Depending on the types of the structures and/or purpose of the studies, such measurements include the quantification of concrete strains, crack geometries, and spalling [12, 14]; masonry cracks [16]; steel buckling and fracture [8]; connection failures [14]; loads and displacements of structural members [13, 14]; and floor accelerations [15]. One limitation of experimental tests, however, lies in the realization that structural models are built via reduced-scale structural components (e.g., beams, columns, slabs, braces, connections, etc.), or substructures (i.e., a small portion of the full structure) for budget considerations. As a result, the experimental findings are bound to limited scopes due to size effects [16]. Conversely, full-scale structural tests can address this concern. However, few studies have been carried out as they are extremely costly and experimentally difficult to perform [17].

*In-situ* post-disaster inspections, on the other hand, can be used in attaining first-hand knowledge of the structural failure(s), hence have been considered effective engineering approaches for structural failure investigations. For instance, Lin et al. [18] surveyed geosynthetic-reinforced soil structures after the 1999 Ji-Ji earthquake in Taiwan and identified the cause of structural failure. Ghasemi et al. [19] visited multiple sites of bridges and tunnels after the Kocaeli earthquake in Turkey and find structural damages during their investigation. Miller [20] summarized the cause of welded connection failures through steel building inspections after the Northridge earthquake.

Thereafter, improvements in structural design have been adopted in revisions of building design codes [21]. A concern of post-disaster inspection, however, is that *in-situ* structure collapse scenes could be disturbed by structural debris. This makes the forensic investigation extremely challenging and time-consuming (consider a progressive building collapse scenario where a primary structural element fails and further results in the failure of adjoining structural elements).

Recently, with the rapid commercialization and technological development of consumer-grade digital cameras and the boom of social network platforms, the events of *in-situ* structural failures can be filmed by cameras and further accessed by the public. The *in-situ* video clips, if available, become a new data source containing valuable spatial-temporal information critical in understanding progressive structural failures. Indeed, the extraction of visual information from *in-situ* video clips can provide direct knowledge of the structural failure behaviors, quantify characteristics of the structural failures, assist in identifying the failure causes, and help first responders to prioritize their efforts in rescue searches.

In this paper, we propose a novel computer vision-based methodology to evaluate the structural collapse of the condominium building in Surfside, Florida. We utilize a video clip of the building collapse that is publicly available on social media, and further apply computer vision algorithms to identify and quantify the falling structural components. The remainder of the paper is structured as follows: Section 2 reviews related work in the literature and overviews the technical contributions of this study; Section 3 demonstrates the research methodology for video processing through a series of vision-based computational algorithms; Section 4 illustrates and discusses preliminary analysis results; Section 5 further evaluate the viability of our method in investigating other building collapse events; Section 6 concludes the study and elaborates the future work.

## 2. Related Work
### 2.1 Vision-Based SHM

Structural health monitoring (SHM) is a proven and effective diagnostic tool in risk assessments and disaster mitigation in civil, mechanical, and aerospace structures. Traditionally, SHM methods rely on sensors to collect measurements of the structure (e.g., vibration signals [22, 23], surface strains [24, 25]). More recently, a paradigm shift points towards the next generation SHM of civil structures, consisting of non-contact, computer vision-based technologies [26]. Compared with contact-based methods, vision-based SHM technologies incorporate the coupled use of digital images and vision algorithms for image processing and pattern recognition, leading to robust, rapid, low-cost methodologies for damage detection and risk mitigation.

To this end, Cha et al. [27] integrated computer vision with deep learning in the context of concrete crack detection in civil structures. In later work, the proposed method was extended to detecting multiple structural deteriorations [28] (i.e., concrete cracks, delamination and corrosions of steel surfaces, and bolt corrections). Narazaki et al. [29] leveraged machine learning and pattern recognition to separate different structural components in bridges using a limited amount of training data, serving as the foundation for an automating visual inspection methodology for a rapid post-disaster response. Yang et al. [30] proposed a novel method using video motion magnification to measure the vibration modes of a three-story reduced-scale structure in the laboratory. The proposed method can identify weakly-excited mode using digital videos. Yeum and Dyke [31] established a vision-based methodology for detecting steel cracks in bridge girders. For a comprehensive overview on this subject, the reader is referred to [32, 33, 34] regarding the recent advances in vision-based SHM.

## 2.2 Video Feature Tracking

Among many vision-based techniques, video feature tracking is one of the fundamental problems in computer vision and has been well studied for a few decades [35]. Nevertheless, video feature tracking had not yet attracted significant attention within the field of vision-based SHM until the last decade. Fukuda et al. [36, 37] were one of the earliest researchers to apply a feature tracking technique for bridge displacement measurements under service loads through video collected by a digital camera. In [36], the proposed method was validated by measuring the vibration of a reduced-scale building frame in the laboratory and a steel girder bridge in the field. Displacements obtained through the vision-based method were found to corroborate well with the results from laser vibrometers and displacement transducers. Similar work that leveraged feature tracking in displacement monitoring has been reported in [38, 39, 40, 41].

Authors previously extended the knowledge of feature tracking from vibration measurement into the area of damage detection of civil structures. In [42], a video feature tracking methodology was established to differentiate the surface motion patterns of cracked and non-cracking regions. As a result, steel fatigue cracks can be consistently identified from video scenes via the utilization of different lighting conditions, surface textures, and camera orientations. Later, this method was improved by the authors by integrating feature tracking with image registration methods for crack detection [43]. Instead of using a video clip, the improved method only relied on two images for finding the breathing behavior of the crack, leading to more robust and accurate crack detection results. Lastly, a similar feature tracking-based method was established in detecting the loosened steel bolts by scanning the entire image scene and identifying the differential visual features provoked by bolt head rotations [44].

To compare feature tracking-based structural vibrations methods against the authors' previous work, one major difference is that vision-based SHM methods for vibration monitoring [36, 37, 38, 39, 40, 41] are based on tracking the visual feature movements of structural members in one or multiple predefined small regions in the image scene. The focus of these studies is to obtain the time-domain vibration measurements of the structural members. In contrast, the authors' methods in [42, 43, 44] leverage feature tracking as a tool to scan the entire image scene, uncovering tiny pixel movements of visual features against two or multiple images that are provoked by subtle differential changes that may not be distinguished by human eyes. As a result, structural damage can be intuitively shown by highlighting the damage locations (e.g., cracks and loosened bolts). Despite the efforts made in the authors' previous work, the potential of feature tracking techniques in solving other engineering problems of vision-based SHM (particularly in structure collapse investigations) has not been studied in the literature and hence remains unknown.

## 2.3 Remote Sensing in Disaster Mitigation

Recently, remote sensing technologies have shown increasing attention in disaster mitigation of civil structures. For instance, Miura et al. [45] trained a convolutional neural network for detecting collapsed buildings from aerial images collected after two earthquakes in Japan. The proposed method has been validated by quantifying the collapsed buildings at a community level. Similar methods were reported by Valentijn et al. [46], Kalantar et al. [47], and Huang et al. [48]. Instead of using aerial images, Yeum et al. [49] and Pan et al. [50] investigated building interior and façade images and developed deep learning algorithms to detect post-disaster structural damages.

The methods discussed above are based on algorithms developed upon post-disaster images. The potential of these methods in applying *in-situ* images or videos that are collected during disaster events has not been explored in these studies. To address this concern, Rigby et al. [51], Diaz [52], and Aouad et al. [53] estimated the yield of the 2020 Beirut explosion based on *in-situ* videos and images downloaded from social media. Results from these studies showed encouraging findings by extracting new information from *in-situ* disaster images and videos, which offered factual evidence to assist the public and stakeholders in border discussions. Muhammad et al. [54, 55] trained deep learning algorithms to detect building fires from *in-situ* surveillance video footage, based on which first responders can control the fire disasters promptly, thus avoiding economic losses. Park et al. [56] proposed a vision-based method to estimate flood depth using images that contain flooded vehicles. Despite the successes reported in these investigations, there remains a dearth of literature regarding the integration of vision-based algorithms in investigating building collapse events.

## 2.4 Citizen Participation and Citizen Sensors

One unique feature of the studies discussed above [51, 52, 53, 54, 55, 56] is that the data are often collected and distributed by the general public. Citizen participation in data collection and distribution can significantly broaden the data sources, through which useful information can be uncovered thus becomes particularly critical for disaster mitigation. For example, the "Did You Feel It?" (DYFI) system [57, 58] developed by the US Geological Survey (USGS) has collected more than five million individual DYFI intensity reports from the general public after seismic events in the past two decades, based on which high-quality macroseismic intensity (MI) maps can be created immediately after earthquake disasters. Similar efforts have been reported in the literature such as iShake [59] and the Quake-Catcher Network (QCN) [60]. Feng et al. [61] investigated smartphone accelerometers as citizen sensors for measuring the vibrations of civil structures in both laboratory and field. The results showed the benefits of these citizen sensors over traditional SHM sensors (considering installation and data acquisition dependencies). Thus, implementing this methodology to a target structure on a routine monitoring basis can be easily achieved.

Despite encouraging research findings through citizen participation, one challenge is how to establish effective computational algorithms for processing citizen-collected SHM data with undesirable data qualities. The general public usually has little or no knowledge about data collection compared with experienced researchers or engineers. This becomes a particular concern in vision-based SHM as the collection of digital images and videos must (generally) follow a predefined data collection protocol. For instance, in studies of vision-based SHM for measuring structural vibrations [36, 37, 38, 39, 40, 41], the video clips were collected by high-fidelity digital cameras under delicate test configurations through enabling reasonable image resolutions, setting up suitable camera shot angles, stabilizing cameras using tripods, adding artificial targets for feature tracking, and/or manipulating suitable lighting conditions. Furthermore, researchers can always recollect videos and images (as many times as they wish) if the data quality is not ideal or the test configuration needs to be improved.

In contrast, the building collapse video adopted in this study (will be discussed in Section 3) was originally extracted from a low-quality surveillance camera and then re-filmed by a hand-held camera. Since neither the owner of the surveillance camera nor the recorder operating the hand-held camera has been trained in professional data collection, the video clip adopted in this study has undesirable qualities in many aspects as itemized below:

- This surveillance camera was intended to capture human activities in the foreground rather than the collapsed building located far from the camera. The Champlain Towers South was not the major consideration factor when the camera view was set up during the camera installation.
- The video quality and image resolution have been negatively affected during the re-filming and redistribution over the internet. As demonstrated in Section 3, the cropped image of the Champlain Towers South is only 641 pixels by 361 pixels, which is much less than the data used in many existing vision-based feature tracking studies in the field.
- The video was collected at night under a dark lighting condition. This environment is substantially different than well-designed experiments, whereby there were no artificial targets installed on the building façade to assist feature tracking.
- The video clip was subjected to a camera movement caused by the operator during re-filming. These further induced uncertainties in feature tracking as the feature movements could be caused by falling structural components and camera movements. An additional strategy is therefore needed to eliminate camera movements.

Based on the former realizations, there exists a clear research gap in the literature to leverage computer vision algorithms to process low-quality SHM data. Although some researchers [62] have investigated the performance of vision sensors when images or videos are highly compressed, these images or videos are collected in the laboratory under controlled environments. The potential of vision-based SHM in extracting information from low-quality videos taken from uncontrolled test setups has not been fully discussed or explored.

**2.5 Contribution**

This study aims to fill the research gaps identified in the preceding subsections. The main technical contributions of this paper are three-fold and are further illustrated below.

First, we extend the knowledge of vision-based feature tracking from applications in structural vibration monitoring into disaster mitigation of civil structures by investigating a building collapse event. Unlike vibration monitoring methods, that focus on tracking time history measurements from small regions in a video scene, the proposed method leverages feature tracking to scan across the entire scene of interest and directly visualize falling structural components by showing their movement directions and magnitudes.

Second, we integrate a series of computer vision algorithms to corroborate new information from an *in-situ* structural collapse video. *In-situ* disaster images or videos contain unique information and may therefore offer new information about the disaster compared with existing investigation methods of building collapse in the literature (i.e., numerical simulations, experimental tests, remote sensing using post-disaster images). Although a growing trend has been reported in the use of *in-situ* videos and images for disaster investigations, to the best knowledge of the authors, there is no literature developing a vision-based methodology for building collapse investigations.

Lastly, we integrate the concept of citizen participation into vision-based SHM to explore the potential of a vision-based methodology through the usage of low-quality data collected under an uncontrolled environment. Many existing vision-based SHM methods rely on delicate test configurations for data collection. Few vision-based SHM studies leverage *in-situ* disaster images or videos collected by untrained citizens. To this end, we investigate surveillance video footage of the building collapse that is distributed from social media corrupted by undesirable qualities. We argue that our investigation may be used to advance the practice of vision-based SHM through the usage of citizen-collected data.

**3. Research Methodology**

Figure 1 illustrates a flow chart of the research methodology for this study with section numbers shown in blue. To begin, Section 3.1 illustrates the detailed protocol for video acquisition by explaining how the video clip adopted for this study is obtained from social media. Section 3.2 elaborates on the technical details of video feature extraction which include two steps: 1) extracting still images (i.e., video frames) from the video clip and subsequentially resizing the video scene by cropping the video frame; and 2) detecting 2D image features from the video frame. Thereafter, Section 3.3 explains the procedure for tracking and pairing detected features against different video frames. Next, Section 3.4 establishes a video stabilization technique that can eliminate the camera movement from the pixel movements of matched feature pairs. In Section 3.5, a novel algorithm is proposed to detect the motion provoked by the falling structural components from the video scene. Lastly, Section 3.6 describes a visualization method to intuitively show the moving directions and magnitudes of falling structural components.

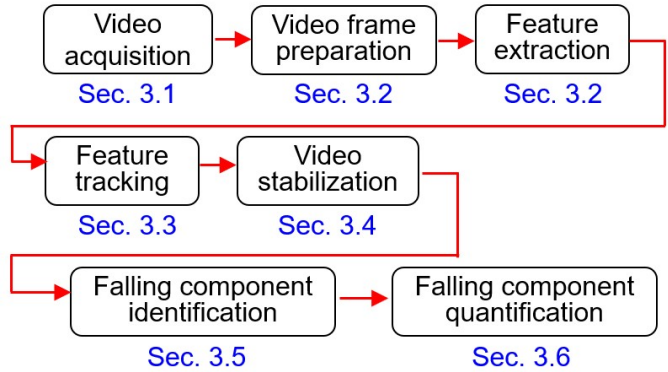

Figure 1. Overview of the research methodology for this study. The numbers shown in blue correspond to article subsections.

### 3.1 Video Acquisition

Figure 2 illustrates the video acquisition procedure. Per ABC Chicago [63], the video was originally captured by a surveillance camera installed in an adjacent condominium building. Then the surveillance video appeared to be displayed on a playback monitor and re-filmed by a hand-held camera. This can be confirmed by observing the small movements (in both translation and scaling) in the video scene caused by the camera movement [64]. Next, the re-filmed video clip was distributed through the internet. Ultimately a copy of this re-filmed video clip was uploaded to YouTube [64] and a digital copy of this video clip was downloaded by the authors through a third-party website YouTube Downloader (YT1s.com). Thereafter, this video clip was adopted as the data source for our study. As of writing this paper, the original surveillance video footage without camera movement has not yet been made publicly available. After saving the video in the MP4 format to a local drive, the video clip was found to have an image resolution of 1920 pixels by 810 pixels with a frame rate of 29.97 frames/sec.

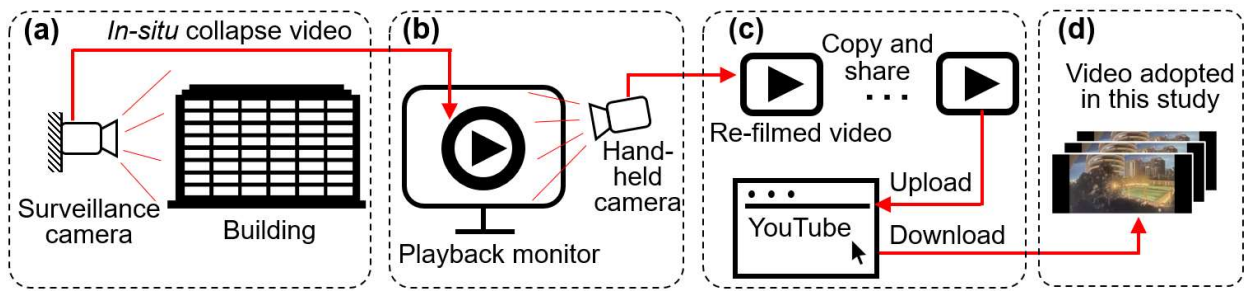

Figure 2. Video acquisition procedure. (a) *In-situ* video collection; (b) video re-filming; (c) video distribution; and (d) video adopted in this study.

### 3.2 Video Feature Extraction

Figure 3 illustrates, for a representative set of frames, the procedure for video feature extraction includes two components: 1) video frame preparation (Figure 3a to c), and 2) feature detection (Figure 3d to g). Below are detailed explanations of each component.

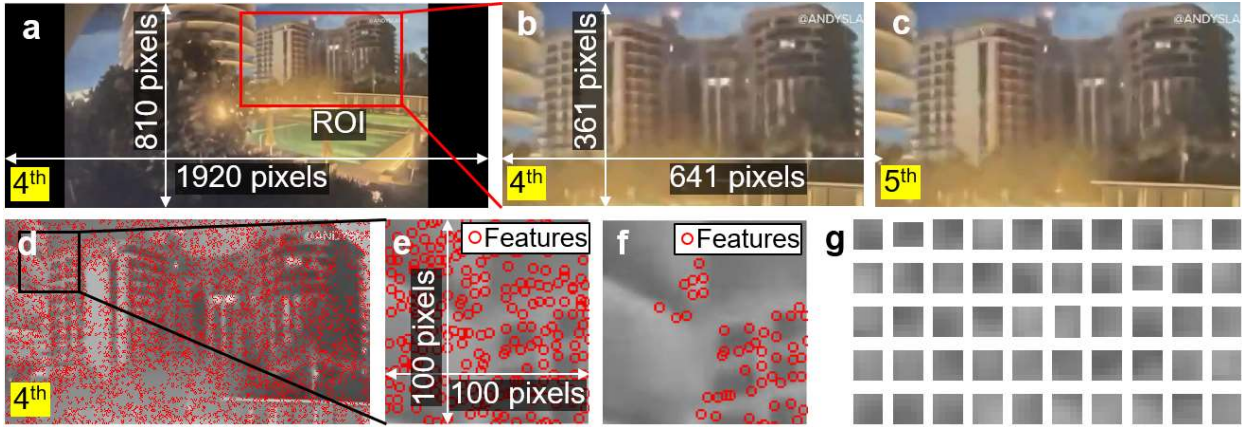

Figure 3. Video feature extraction. (a) The original 4th video frame; (b) the cropped 4th video frame under ROI; (c) the 5th video frame under ROI; (d) Shi-Tomasi features of the 4th video frame under ROI; (e) the blow-up view with all Shi-Tomasi features; (f) the blow-up view of the 50 most distinguishable Shi-Tomasi features; and (g) image patches of the 50 most distinguishable Shi-Tomasi features.

Video frame preparation starts with extracting a total of 723 video frames (i.e., still images) from the video clip. To uncover the differential changes of visual features against different video frames, two adjacent video frames are selected. For illustration, here we adopt the 4th and 5th frames at the beginning of the video clip in this section while video frames from other time instants will be illustrated in Section 4. The 4th video frame is first shown in Figure 3a and is then cropped using a region of interest (ROI) (see the red box in the figure). The boundary of the ROI is manually defined to quantify the video scene of the Champlain Towers South. The ROI has a dimensionality of 641 pixels by 361 pixels and the 4th frame after applying the ROI is shown in Figure 3. The same ROI is applied to the 5th video frame and the cropped image is shown in Figure 3c. Applying ROIs for cropping video frames can remove unnecessary objects (i.e., the adjacent building, the playground, plants, light posts) from the video scene such that feature detection (explained later in this section) can be applied to the scene of the Champlain Towers South only.

In terms of feature detection, the ROI of the 4th video frame (Figure 3b) is first converted from RGB channels into a greyscale image (Figure 3d). Then, Shi-Tomasi features [65] (red circles in Figure 3d) are extracted in the ROI. The feature point describes a location where its adjacent area has unique intensity changes that are invariant against image translation, rotation, and scaling; hence the same feature point can be consistently found and paired in different video frames. In addition to Shi-Tomasi features, other common 2D image features include features from accelerated segment test (FAST) [66], Harris-Stephens [67], binary robust invariant scalable keypoints (BRISK) [68] and speeded up robust features (SURF) [69]. The comparison of these features in image matching in vision-based SHM problems is investigated in [44]. To better demonstrate the concept of feature detection, Figure 3e shows all the detected Shi-Tomasi features within an area of 100 pixels by 100 pixels under a blow-up view. Figure 3f further lists the 50 most distinguishable Shi-Tomasi features, and their adjacent areas are plotted as small image patches (5 pixels by 5 pixels) in Figure 3g.

## 3.3 Video Feature Tracking

Once the features are detected from the 4$^{th}$ and 5$^{th}$ frames, a feature tracker is deployed to pair features between these two frames. To this end, we apply the Kanade-Lucas-Tomasi (KLT) tracker [70, 71] to the ROIs of the 4$^{th}$ and 5$^{th}$ video frames. The KLT tracker obtains the information of feature points in the 4$^{th}$ video frame (Figure 3d) and actively scans the same ROI of the 5$^{th}$ video frame (in greyscale) to match feature points with similar intensity changes. Figure 4a shows all matched feature pairs as a result of KLT tracking, where red circles are matched features in the 4$^{th}$ video frame; and green crosses are the matched features in the 5$^{th}$ frame.

To better illustrate the feature tracking details, five areas are selected in Figure 4a where A1 is on the balcony in a foreground building; A2, A3, and A4 are different locations from the building complex of the Champlain Towers South; and A5 is at the roof of a foreground structure. Their corresponding blow-up views are shown in Figure 4b to f. As can be seen from these figures, the small movements of matched feature pairs can be found in both the Champlain Towers South and the foreground structures (A1, A2, A3, and A5). This is likely to be driven by the hand-held camera movements during the re-film process. On the other hand, pixel movements are large in A4 of the Champlain Towers South (Figure 4e). Such movements are likely to be provoked by a combination of hand-held camera movements and the motions of falling structural components. The verifications of these hypotheses will be demonstrated in Section 3.4.

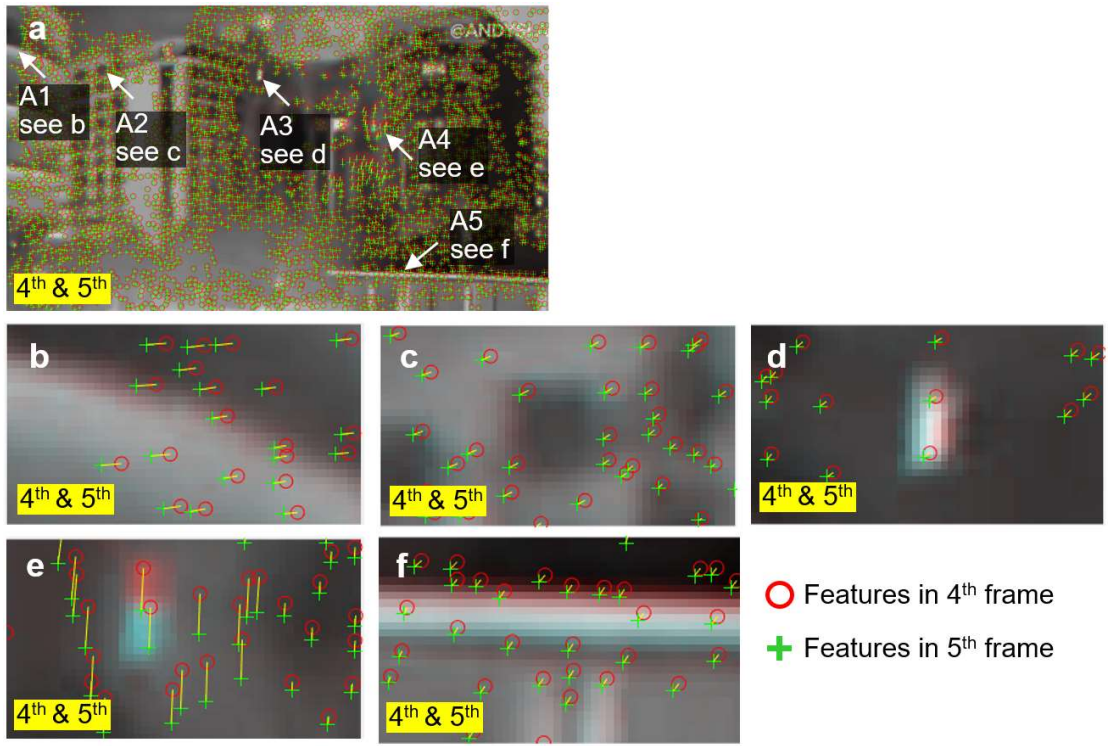

Figure 4. Video feature tracking. (a) Matched feature pairs between the 4$^{th}$ and 5$^{th}$ video frames under ROI defined in Figure 3a, from which five areas (A1 to A5) are selected; blow-up views of A1 to A5 are shown in (b) to (f). Red circles are matched features in the 4$^{th}$ frame; green crosses are matched features in the 5$^{th}$ frame.

## 3.4 Video Stabilization

To further improve the results of feature tracking, we herein apply a video stabilization method as illustrated in Figure 5. To explain, suppose two video frames (Frames 1 and 2) are extracted from a video clip taken by a hand-held camera about a still scene. Due to the camera movement provoked by hands, Frame 2 is under a different image coordinate system than Frame 1, being subjected to an image (out-of-plane) rotation, translation, and scale (caused by camera zoom-in). The video stabilization protocol starts by detecting features from both video frames. Then, these features are paired together using the video feature tracking methodology explained in Section 3.2. Thereafter, a geometric transformation matrix is established [72] using matched feature pairs. The matrix describes the geometric relationship of image coordinate systems of two video frames. Next, we treat Frame 1 as the reference image and apply the geometric transformation matrix to Frame 2. As a result, Frame 2 can be recovered to the same coordinate system as Frame 1, and the camera movement induced between two frames hence is eliminated.

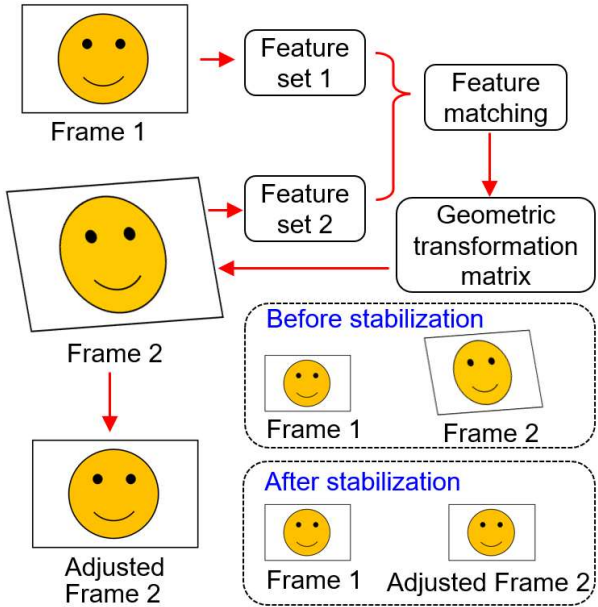

Figure 5. Video stabilization flowchart

Figure 6 illustrates the most common types of 2D geometric transformations that include translation, rigid, similarity, affine, and projective [73]. The projective geometric transformation is adopted in this study that can estimate the out-of-plane tilted movements caused by the hand-held camera. To explain, denote $x_1$ and $y_1$ are coordinates of arbitrary 2D points on Frame 1 in Figure 5; and $x_2$ and $y_2$ are coordinates of arbitrary 2D points on Frame 2. The geometric relation between Frames 1 and 2 then can be established as:

$$\begin{bmatrix} x_2 \\ y_2 \\ 1 \end{bmatrix} = \boldsymbol{H} \begin{bmatrix} x_1 \\ y_1 \\ 1 \end{bmatrix} = \begin{bmatrix} a & b & c \\ d & e & f \\ g & h & 1 \end{bmatrix} \begin{bmatrix} x_1 \\ y_1 \\ 1 \end{bmatrix} \tag{1}$$

where $\boldsymbol{H}$ is the geometric transformation matrix. Using matched feature pairs between Frames 1 and 2 (i.e., correspondences), $\boldsymbol{H}$ can be solved and further applied to Image 2 for recovering adjusted Frame 2. For more details of 2D geometric transformations, the reader is referred to [74, 75].

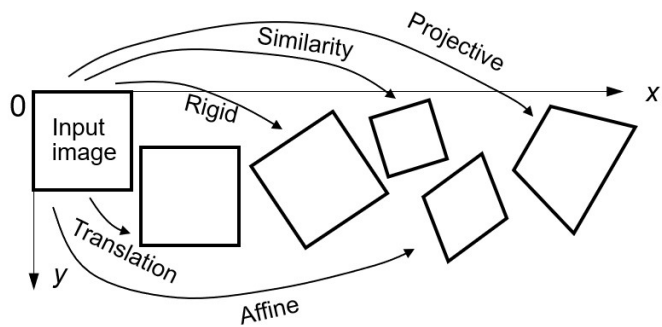

Figure 6. Different types of 2D geometric transformations

The results after applying the video stabilization method to the building collapse video clip are shown in Figure 7. To this end, the geometric transformation matrix $\boldsymbol{H}$ is first established through the matched feature pairs between the 4$^{th}$ and 5$^{th}$ video frames in Figure 4a. $\boldsymbol{H}$ is calculated as:

$$\boldsymbol{H} = \begin{bmatrix} 0.99 & 0 & 0 \\ 0 & 0.99 & 0 \\ 1.85 & -0.24 & 1 \end{bmatrix}$$

Then, $\boldsymbol{H}$ is applied to the 5$^{th}$ frame to recover it to the same image coordinate system as the 4$^{th}$ frame. Figure 7a shows the matched feature pairs between the 4$^{th}$ frame and the adjusted (i.e., recovered) 5$^{th}$ frame after video stabilization, where red circles are matched features in the 4$^{th}$ video frame; and yellow crosses are the matched features in the 5$^{th}$ frame.

To compare feature matching results before and after video stabilization, five same areas (A1 to A5) are selected in Figure 7a. As can be found by comparing A1, A2, A3, and A4 in Figure 4 against Figure 7, the small pixel movements that occur in these areas can be dramatically reduced after applying video stabilization. Hence the effect of camera movements can be successfully eliminated in these areas.

On the other hand, large pixel movements can be found in A4 where structural components are falling in both Figure 4 against Figure 7. To investigate the feature movements in this area, we denoted $\boldsymbol{a}$ as the feature movement of the 4$^{th}$ and 5$^{th}$ video frames before video stabilization (see Figure 8). Then $\boldsymbol{a}$ can be decomposed into the summation of two vectors: movement caused by falling structural components $\boldsymbol{b}$ and camera movement $\boldsymbol{c}$. After video stabilization is applied to the 5$^{th}$ video frame, the camera movement $\boldsymbol{c}$ is eliminated. The movement that remained between the feature in the 4$^{th}$ video frame (red circle) and the feature in the adjusted 5$^{th}$ video frame (yellow cross) is the structural movement $\boldsymbol{b}$, hence shall be considered as the movement provoked by the falling structural components. This phenomenon can be also verified by comparing blow-up views of A4 in Figure 4e against Figure 7e.

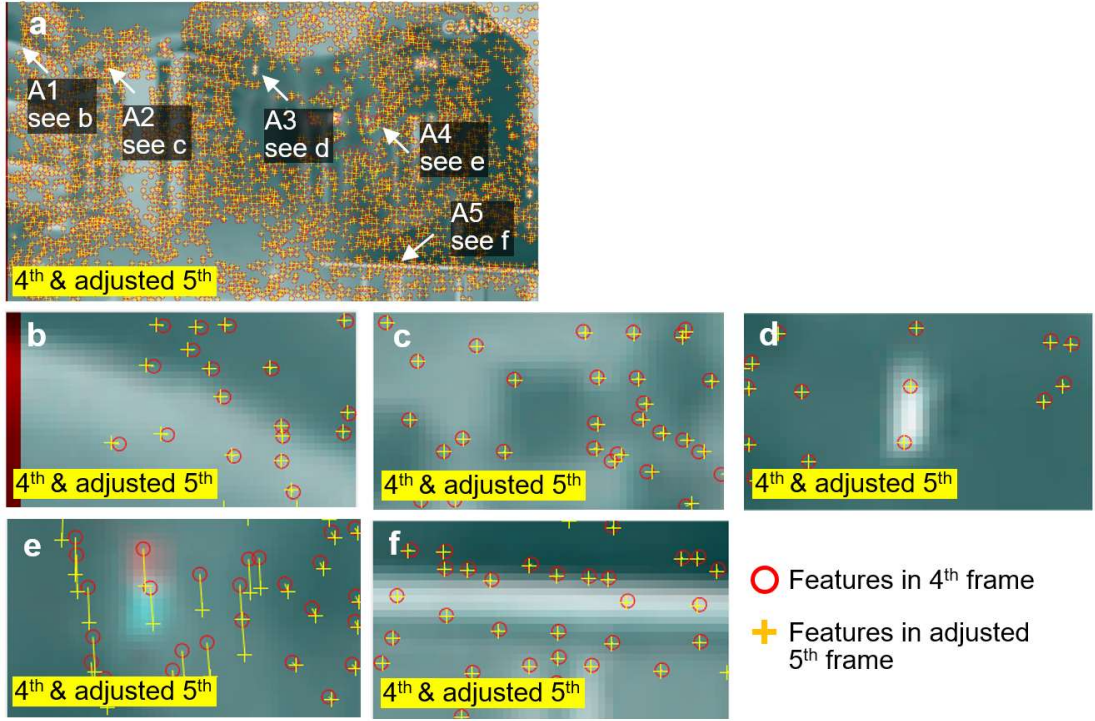

Figure 7. Video stabilization. (a) Matched feature pairs between the 4$^{th}$ and adjusted 5$^{th}$ video frames after video stabilization. Five areas (A1 to A5) same as Figure 4a are selected in (a) and their blow-up views of A1 to A5 are shown in (b) to (f). Red circles are matched features in the 4$^{th}$ frame; yellow crosses are matched features in the 5$^{th}$ frame.

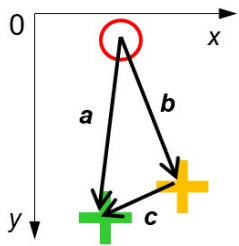

*a*: Pixel distance before stabilization
*b*: Pixel distance after stabilization
  (structural movement)
*c*: Camera movement

○ Features in 4$^{th}$ frame
+ Features in 5$^{th}$ frame before stabilization
+ Features in 5$^{th}$ frame after stabilization

Figure 8. The decomposition of pixel movements of matched feature pairs in A4.

## 3.5 Identification of the Falling Components

Figure 9a provides a schematic diagram to illustrate a methodology for finding the falling structural components during the building collapse. To explain, we treat the pixel movements of matched feature pairs after video stabilization (Figure 7a) as robust indicators for identifying structural motion. The feature points located in the falling components usually have large movements (see Figure 7e after video stabilization). In contrast, the feature points around intact areas of the building complex are subjected to extremely small movements (see Figure 7c). By applying a cutoff threshold *ts* and hiding these features that are below *ts*, the remainder of the features can be highlighted in the video scene to identify the falling components.

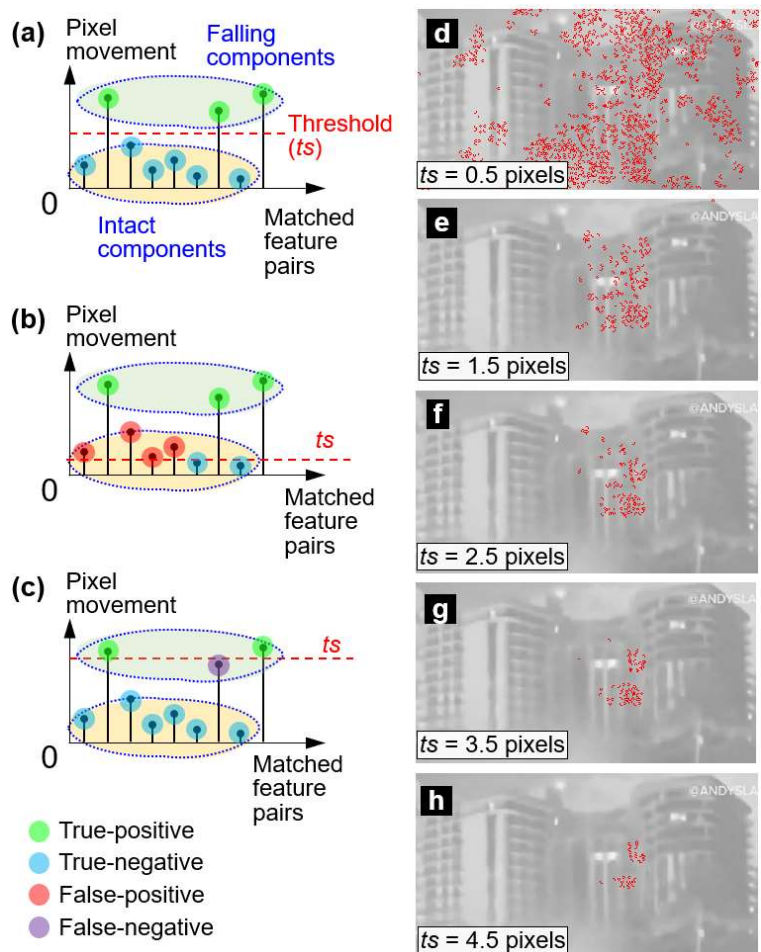

Figure 9. Methodology for identifying the falling structural components with different cutoff thresholds *ts* are shown in (a), (b), and (c). Results after applying different *ts* are shown in (d) to (h), respectively. The background images from (d) to (h) are taken from the 4$^{th}$ video frame with image brightness adjusted for the illustration purpose.

The cutoff threshold *ts* is highly related to the moving magnitudes of the falling components. For instance, a building section that just starts to collapse would yield much smaller pixel movements than the movements caused by these components that fall for a while with large velocities. Sometimes, the combination of these two situations may appear in one scene simultaneously. Since the moving magnitudes of structural components are not usually known as *a priori*, challenges exist in predetermining a suitable value of *ts*. As shown in Figure 9b, a low *ts* could produce false-positive results by incorrectly identifying features with extremely small pixel movements in the intact areas. On the other hand, a large *ts* could accidentally cut off features with small pixel movements caused by the falling components, leading to false-negative detection results as shown in Figure 9c.

To address this concern, here we adopt an iteration procedure by applying different values of *ts* and then select a suitable *ts* based on observations. Figure 9d to h show the matched feature pairs after applying different *ts*. For illustration purposes, only features from the 4$^{th}$ video frame are shown. As can be found in the figures, a low *ts* yield false-positive identification results (see Figure 9d); while a high *ts* can produce filter out true positive results with small pixel movements (see Figure 9h) hence less falling components are detected. To balance the trade-off, *ts* is selected as 3.5 pixels for the rest of the video processing in Section 3.

### 3.6 Quantification of the Falling Components

Figure 10a to c demonstrate the method for quantifying the falling components. To visualize the falling structural components, matched feature pairs between the 4$^{th}$ and 5$^{th}$ video frames are replaced by arrows. The arrows start from the matched feature point in the 4$^{th}$ frame and point to the matched feature point in the 5$^{th}$ frame. The magnitudes of the arrows are the distances of image coordinates between matched feature pairs (Figure 10b). Because feature movements are usually very small (e.g., only a few pixels), here we amplify the arrow lengths by multiplying a unified scaling factor as shown in Figure 10c, enabling a more intuitive way of viewing the result graphically. Next, we apply this method to all matched feature pairs in Figure 9e after the cutoff threshold (*ts* = 3.5 pixels) is applied. The results are shown in Figure 10d and e. The benefit of using arrows is two-fold: 1) they indicate the directions structural components moved; and 2) the magnitudes of arrows qualitatively represent the distances of matched feature pairs moved within a one video frame interval (i.e., the arrow magnitude indicates the velocities of the falling component).

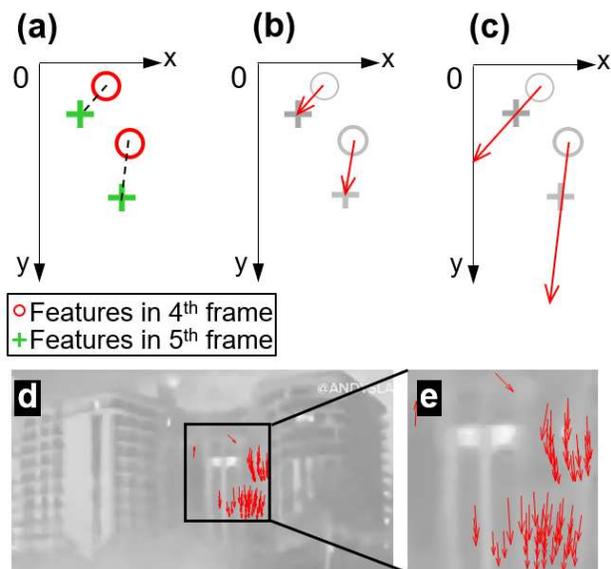

Figure 10. (a) Two matched feature pairs between the 4$^{th}$ and 5$^{th}$ video frames; (b) the matched feature pairs are replaced by two arrows; (c) two arrows after applying a scaling factor; (d) arrows created in 4$^{th}$ video frame; (e) the blow-up view of the arrows. The background images in d and e are taken from the 4$^{th}$ video frame with its image brightness adjusted for the illustration purpose.

## 4. Preliminary Results and Discussions

We identify five building collapse instants across the video clip that depict critical moments of building collapse. For each instant, two adjacent video frames are selected for video processing using MATLAB [76]. As shown in Table 1, the corresponding timelines of selected video frames are calculated using a frame rate of 29.97 frames/sec. The selected video frames and analysis results are summarized in Figure 11.

Table 1: Matrix for video processing

|  | Video frame | Timeline | Result |
|---|---|---|---|
| Instant 1 | 4th and 5th | 0.13 and 0.17 seconds | Figure 11c |
| Instant 2 | 17th and 18th | 0.57 and 0.6 seconds | Figure 11g |
| Instant 3 | 30th and 31st | 1 and 1.03 seconds | Figure 11k |
| Instant 4 | 52nd and 53rd | 1.73 and 1.77 seconds | Figure 11o |
| Instant 5 | 309th and 310th | 10.3 and 10.33 seconds | Figure 11s |

As can be found in Figure 11, the middle section of the building complex starts falling at the beginning of this video clip (see the first, second, and third columns in Figure 11). Consistently, the red arrows signify the falling structural components and further quantify their moving directions and magnitudes. Next, additional components in the middle section fall (see the fourth column). It is worth noting that the camera view is partially obstructed by particulate matter made by the prior collapsing building debris. Thereafter, the right section of the building complex falls as can be seen in the last column of the figure.

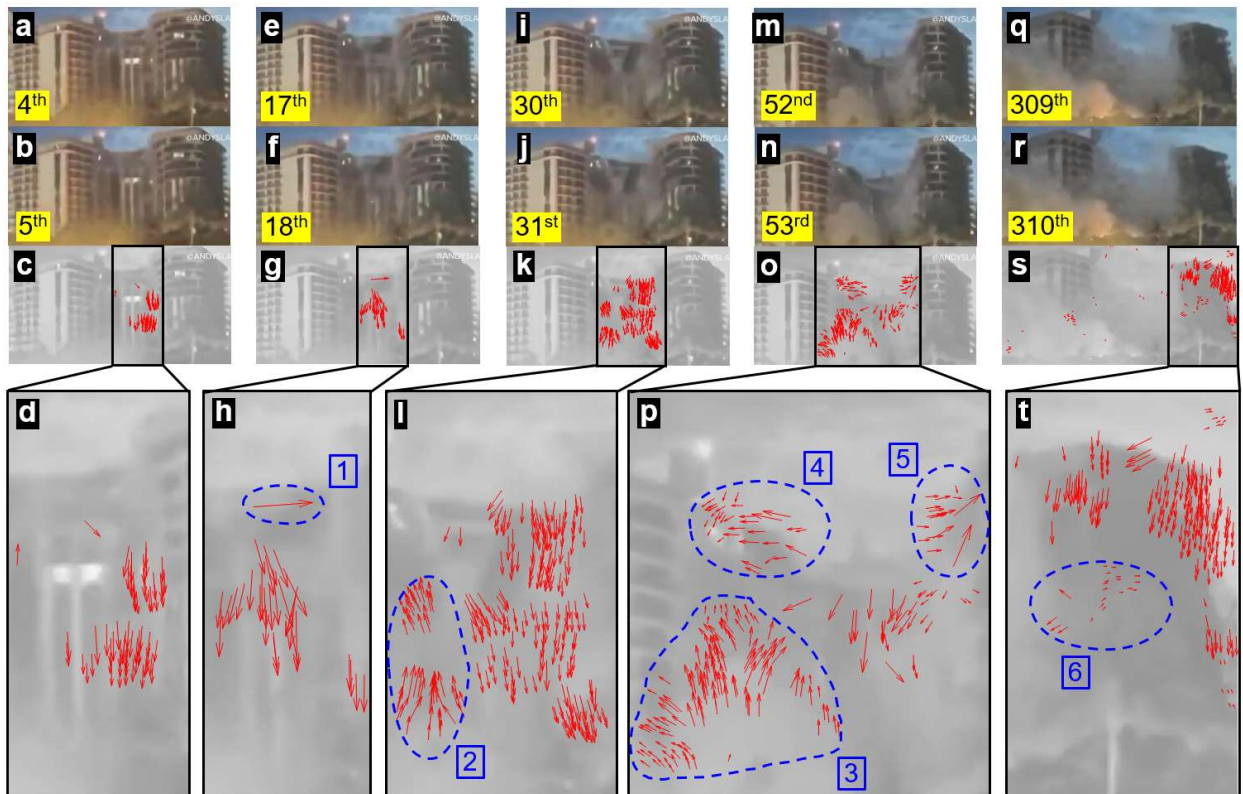

Figure 11. Video processing results where Columns 1 to 5 are results for Instants 1 to 5 from the video clip. The first and second rows are the selected video frames for the analysis; the third row shows the video

analysis results with arrows indicting the moving directions and magnitudes of the falling structural components; the last row shows blow-up views. The labels at the lower-left corners of the images at the first and second rows are the video frame numbers. The brightness of images in the third and last rows are adjusted for the purpose of illustration.

Several areas of the images in the last row in Figure 11 are highlighted by blue dotted circles. In dotted circle 1, the horizontal arrow may be caused by an incorrect KLT matching and hence may be considered as the outlier. In dotted circles 2, 3, and 6, arrows are plotted in the upward directions due to movements of particulate matter (dust and/or smoke). This can be further confirmed by inspecting the particulates in video frames (e.g., Figure 11i and q). In dotted circle 4, arrows depict the horizontal movements of the structural components. This is herein considered the correct detection as the local structural components may be initially attached to the left building section through reinforcing steel bars in the slabs, which prevent the components from moving downward. In dotted circle 5, horizontal and upward arrows are plotted. This may be caused by both the horizontal movements of the structural components and the upward movements of the particulates.

Overall, the results in Figure 11 successfully deliver new knowledge that is beyond human interrelation by inspecting the video through one's eyes. Take the results from the $4^{th}$ and $5^{th}$ frames (first column in the figure) as an example. Our proposed method can effectively scan the entire video scene to rapidly identify the locations of falling structural components, and intuitively quantify the directions and magnitudes of these components through arrows. Notice that such information is solely based on tracking differential features between two video frames. On the other hand, visual inspection based on human eyes may face difficulties in achieving a similar level of understanding/recognition.

Also, notice that success validation of our method in this study is based on a low-quality video clip with very limited spatial information in comparison to, for example, high fidelity laboratory data [36, 37, 38, 39, 40, 41]. Namely, because true timestamps are not available, the video playback speed (i.e., timeline) in the video clip adopted in this study may vary in comparison to the actual event time. Secondly, the pixel resolution and frame rate of original *in-situ* surveillance video footage remain unknown. Lastly, the video clip likely does not capture the initiation of the collapse event (i.e., the surveillance camera may be triggered by the motion of building collapse, leading to a delay of camera filming). Notice that our approach solely relies on extracting information from existing video frames. Resultingly, potential losses of initial video frames would not affect the results of this study.

Because particulate matter always appears on the building collapse scene immediately after the first batch of structural components reaches the ground, it is recommended to apply the proposed method at the very beginning of the collapse event (e.g., $4^{th}$ and $5^{th}$ frames) or the beginning of the follow-up collapse event (e.g., $309^{th}$ and $310^{th}$ frames). In both cases, the camera view of the building has not been (fully) covered by the particulate matter yet. As shown in Figure 11, the video footage between $53^{rd}$ and $309^{th}$ (the fourth and last columns) may not be the ideal dataset for processing our method of identifying the structural movements of the middle section of the building complex. This is since the particulate matter has already spread across and obstructed the view of the middle section. Nevertheless, using our method in processing initial video frames is still essential for understanding the cause of structural failure. Additionally, the use of advanced regularization methods, especially sparse ($L_1$ or Total Variation [77]) approaches may improve video information via the elimination of image artifacts (e.g., particulate matter) and will be investigated in the future.

Also shown in Figure 11, the matched feature pairs may not be evenly distributed across the entire falling structural components, leading to a scattered pattern of arrow distributions. This is due to the difficulty of the KLT tracker in matching the feature points from different video frames. Factors that can negatively affect the KLT matching capability include the low quality of the surveillance camera, potential video quality losses during the procedures of video re-filming, distributing, and downloading, and the dark lighting condition at night.

In addition to the former technical discussion, it is also important to recognize recent advances, progress, and thrust potential made available by large disaster investigation collaborations. For example, capabilities including advanced imaging, Unmanned Aerial System (UAS) data collection, laser, and lidar scanning, originating from natural disaster response and reconnaissance needs (cf. Natural Hazards Engineering Research Infrastructure (NHERI) as a recent exemplar [78]), offer a large suite of capabilities affording significant capacity for post-collapse forensics. In conjunction with image processing approaches, as described herein, such efforts can offer enriched joint information for detailed forensics and analysis imperative for immediate response and detailed long-term study.

## 5. Viability in Investigating Other Structure Collapse Events

To evaluate the viability of our proposed method in investigating other building collapse events, here we adopt another publicly available video clip [79] which compiles demolitions of a few hotel buildings through implosions in the past decades in Las Vegas, Nevada of the United States. The video was downloaded from YouTube and was found in a resolution of 1920 pixels by 1080 pixels with a frame rate of 60 frames/sec after saving in a local drive. Then four demolition events were selected from this video clip with different lighting conditions (e.g., daytime, night) and structure types (e.g., tower, building). The selected demolished hotels are the LandMark (demolished in 1995), the Sands Hotel and Casino (demolished in 1996), the Desert Inn (demolished in 2004), and the Castaways Hotel and Casino (demolished in 2006).

Figure 12 to Figure 15 show results for tracking structural motions during all four hotel demolitions. For the illustration purpose, here we use the demolition of the LandMark (Figure 12) to illustrate key information in the figures. To explain, we first select two adjacent video frames (denote $1^{st}$ and $2^{nd}$ frames) from the beginning of this building demolition event. In the $1^{st}$ frame, an ROI (see the red box in Figure 12a) is marked from the video scene. The location of the implosion (LOI) with green arrow(s) is also added to the figure to indicate where the implosion occurs in the structure at the beginning of the demolition. Figure 12b and c are blow-up views from the selected ROI under different video frames, and the final processing result of the selected ROI is shown in Figure 12e.

Since the ground truth measurements of structural motion are not available, here we propose a new criterion to evaluate the accuracy of our method by comparing our method (Figure 12e) with the absolute intensity difference between two frames (Figure 12d). The absolute intensity difference is calculated by extracting the absolute intensity difference between Figure 12b and c. As shown in Figure 12c, if pixel intensities between Figure 12b and c are exactly matched, the corresponding image pixels are illustrated as black (0 intensity); while intensities of unmatched pixels are shown in grey to white across a range of 1 to 255 intensity values, depending on the level of discrepancy. As can be seen in Figure 12e, the locations of red arrows are in about the same locations as the bright areas in Figure 12d with higher intensity values. These areas are provoked by the subtle structural motion between video frames caused by the implosion indicated in Figure 12a. Our method is superior to the absolute difference method as it does not only identify the locations where structural motions occur, but also quantifies the directions and magnitudes of the falling structural members.

Results across Figure 12 to Figure 15 indicate that our method can reliably quantify structural motions from different building demolitions, regardless of the variations in lighting conditions, building types, implosion locations, sizes of ROI, original video qualities, and building surface textures. Compared with human visual inspections, the proposed vision-based methodology can bring new knowledge of the building collapses by tracking differential features from video frames.

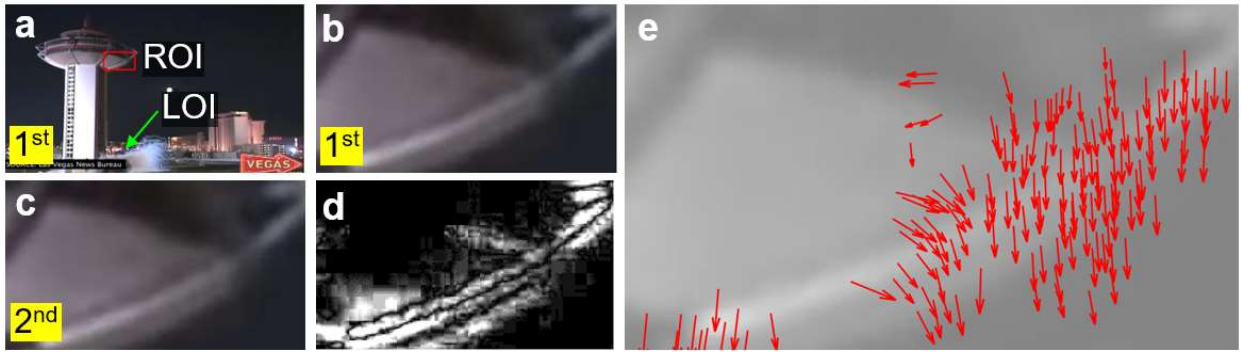

Figure 12. Video processing results of demolition of the LandMark. (a) shows the video scene; (b) and (c) are blow-up views of the ROI from two adjacent views; (d) is the absolute greyscale intensity difference between (b) and (c); and (e) is the result processed by our method under the ROI. The labels at the lower-left corners in (a) to (c) are the video frame numbers. The image brightness in (e) is adjusted for illustration. ROI and LOE refer to the region of interest and location of implosion, respectively.

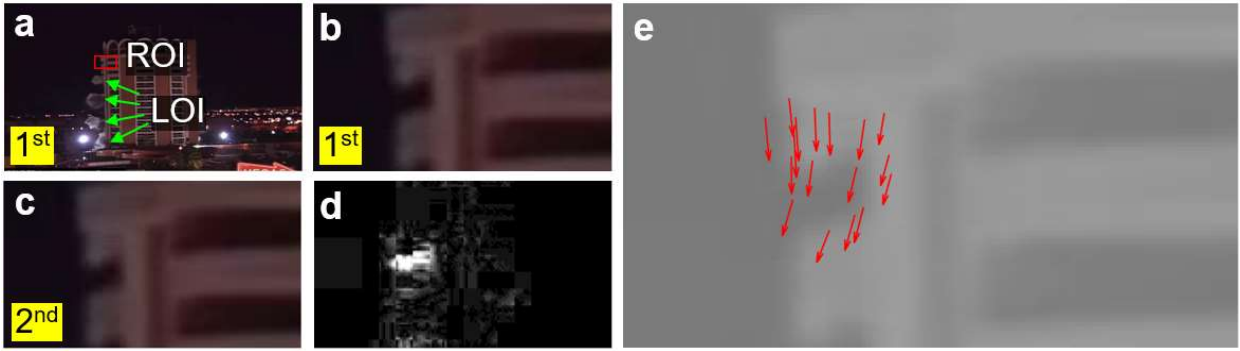

Figure 13. Video processing results of demolition of the Sands Hotel and Casino. (a) shows the video scene; (b) and (c) are blow-up views of the ROI from two adjacent views; (d) is the absolute greyscale intensity difference between (b) and (c); and (e) is the result processed by our method under the ROI. The labels at the lower-left corners in (a) to (c) are the video frame numbers. The image brightness in (e) is adjusted for illustration. ROI and LOE refer to the region of interest and location of implosion, respectively.

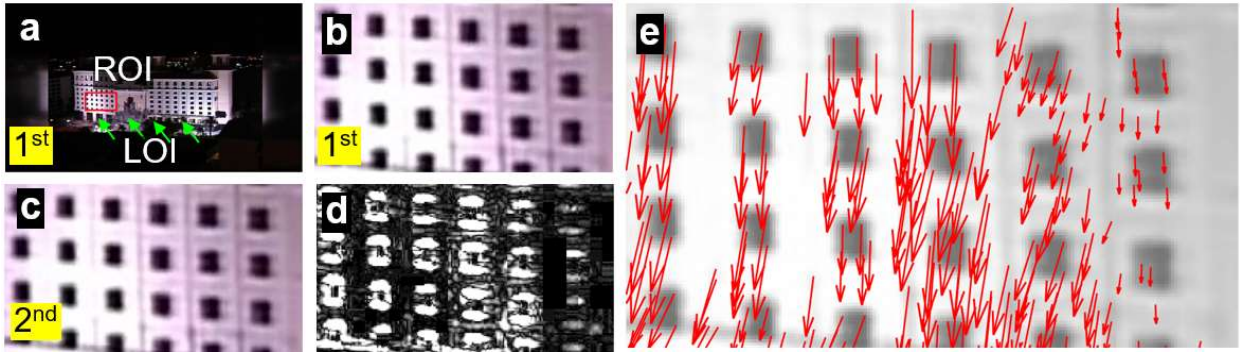

Figure 14. Video processing results of demolition of the Desert Inn. (a) shows the video scene; (b) and (c) are blow-up views of the ROI from two adjacent views; (d) is the absolute greyscale intensity difference between (b) and (c); and (e) is the result processed by our method under the ROI. The labels at the lower-left corners in (a) to (c) are the video frame numbers. The image brightness in (e) is adjusted for illustration. ROI and LOE refer to the region of interest and location of implosion, respectively.

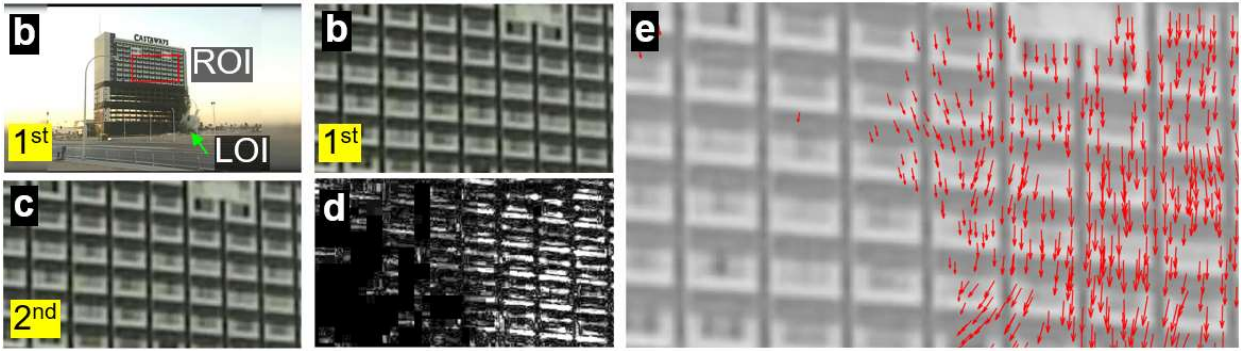

Figure 15. Video processing results of demolition of the Castaways Hotel and Casino. (a) shows the video scene; (b) and (c) are blow-up views of the ROI from two adjacent views; (d) is the absolute greyscale intensity difference between (b) and (c); and (e) is the result processed by our method under the ROI. The labels at the lower-left corners in (a) to (c) are the video frame numbers. The image brightness in (e) is adjusted for illustration. ROI and LOE refer to the region of interest and location of implosion, respectively.

## 6. Conclusions and Future Work

The investigation of a structural collapse requires joint efforts from structural engineers, government agencies, law enforcement, and other stakeholders. The investigation team needs to examine every piece of evidence related to the event, among which the *in-situ* collapse video clip, if available, is one of the most essential pieces of evidence. Traditionally, structural collapse video clip/clips is/are usually visually inspected [80]. In this paper, we present a novel video processing methodology to extract new information from the collapse video clip by showing the directions and magnitudes of the movements of the falling components. Our method aims to help investigators in obtaining direct knowledge of the behaviors and mechanisms of the building collapse.

The proposed method can be implemented in different numerical platforms and the computational time for generating results from two adjacent frames in this study is about seven seconds using a workstation with a 2.9 GHz CPU and a 32 GB RAM. These flexibilities allow the proposed method to be implemented in rescue searching shortly after the building collapse. As shown in the last row of Figure 11, the proposed method can provide a factual interpretation of the sequence of the collapsing components (e.g., which portion of the building collapses first and hence shall be at the bottom of the debris). This information could help first responders to prioritize their efforts in the rescue mission after the disaster.

Recently, with the rapid commercialization and technological development of consumer-grade digital cameras, structure collapses could be filmed by a variety of means. In addition to surveillance cameras, other video collection tools may include smartphone cameras, dash cameras on passing vehicles, body cameras from onsite construction workers, and drones operated by rescue teams. These new citizen-collected data bring great potential for integrating vision-based SHM methods to generate new knowledge for disaster mitigation. Further work will focus on investigating the viability of this video processing methodology to border applications such as quantification of the global structural behavior of the collapsed building.

**Acknowledgment**

The authors convey condolences to the families affected by the building collapse and want to thank those who collected the (surveillance) video footage of the building collapses in this study; and those who shared and distributed the video clip over social media. The authors also want to thank William Ambrose for his fruitful discussions; Ariana Baker and Eric Resnis for their help; and anonymous reviewers for improving the quality of this paper.

**Funding**

The first author is partially supported by the new faculty start-up research fund from the Gupta College of Science at Coastal Carolina University. The second author would like to acknowledge the support via Engineering and Physical Sciences Research Council Project EP/V007025/1.

However, any opinions, findings, and conclusions, or recommendations expressed in this study are those of the authors and do not necessarily reflect the views of the funders.

**Conflicts of Interest**

The authors declare no conflict of interest.

**Data Availability Statement**

The data that support the findings of this study are openly available in the internet sources referred to in the paper.